\documentclass{article}

\usepackage[small]{caption}
\usepackage{emnlp2017}
\usepackage{times}
\usepackage{latexsym}
\usepackage{tablefootnote}
\usepackage{url}
\usepackage{color}
\usepackage{float}
\usepackage{tabularx}
\usepackage{subfig}
\usepackage{multirow}
\usepackage{latexsym}
\usepackage{amsmath}
\usepackage{amssymb}
\usepackage{multirow}

\usepackage{pgfplots}
\usepackage{ulem}

\emnlpfinalcopy 

\DeclareMathOperator*{\argmax}{arg\,max}
\def\x{\mathbf{x}}

\def\bz{\mathbf{z}}

\def\g{\textbf{g}}
\def\h{\textbf{h}}

\def\br{\mathbf{r}}
\def\p{\mathbf{p}}

\def\bb{\mathbf{b}}

\def\bW{\mathbf{W}}

\def\bA{\mathbf{A}}

\def\R{\mathbf{R}}
\def\T{\mathcal{T}}

\title{A Feature-Enriched Neural Model for Joint Chinese Word Segmentation and Part-of-Speech Tagging}
\author{Xinchi Chen, Xipeng Qiu\thanks{{ }{ }Corresponding author.}, Xuanjing Huang\\
 Shanghai Key Laboratory of Intelligent Information Processing, Fudan University\\
School of Computer Science, Fudan University\\
825 Zhangheng Road, Shanghai, China\\
\{xinchichen13,xpqiu,xjhuang\}@fudan.edu.cn}

\date{}

\begin{document}

\maketitle

\begin{abstract}
Recently, neural network models for natural language processing tasks have been increasingly focused on for their ability of alleviating the burden of manual feature engineering. However, the previous neural models cannot extract the complicated feature compositions as the traditional methods with discrete features. In this work, we propose a feature-enriched neural model for joint Chinese word segmentation and part-of-speech tagging task. Specifically, to simulate the feature templates of traditional discrete feature based models, we use different filters to model the complex compositional features with convolutional and pooling layer, and then utilize long distance dependency information with recurrent layer. Experimental results on five different datasets show the effectiveness of our proposed model.
\end{abstract}

\section{Introduction}

\noindent Chinese word segmentation and part-of-speech (POS) tagging are two core and fundamental tasks in Chinese natural language processing (NLP). The state-of-the-art approaches are based on joint segmentation and tagging (S\&T) model, which can be regarded as character based sequence labeling task. The joint model can alleviate the error propagation problem of pipeline models.

Previously, the traditional hand-crafted feature based models have achieved great success on joint S\&T task \cite{Jiang:2008,kruengkrai2009error,Qian:2010,zhang2008joint,Zhang:2010}. Despite of their success, their performances are easily affected by following two limitations.

The first is \textbf{model complexity}. Since the decoding space of joint S\&T task is relatively large, the traditional models often rely on millions of discrete features. Therefore, the efficiency of joint S\&T models is rather low. Moreover, these models suffer from data sparsity. Recently, some neural models \cite{huang2015bidirectional,chen2015gated,ma2016end} are proposed to reduce the efforts of feature engineering and the model complexity. However, these neural models just concatenate the embeddings of the context characters, and feed them into neural network. Since the concatenation operation is relatively simple, it is difficult to model the complicated features as the traditional discrete feature based models. Although the complicated interactions of inputs can be modeled by the deep neural network, the previous neural models show that the deep model cannot outperform the one with a single non-linear model.

\begin{figure}[t]
 \centering
 \includegraphics[width=0.48\textwidth]{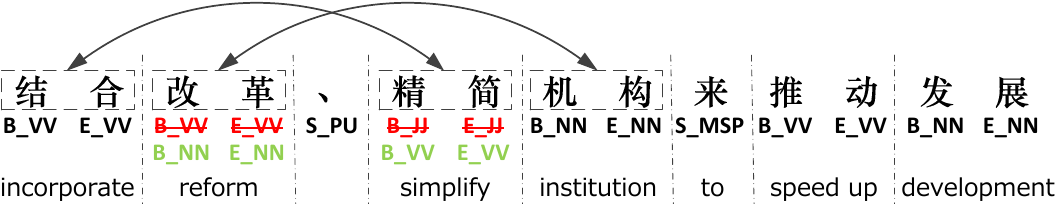}
 \caption{An example. CRF makes mistakes on words ``reform'' and ``simplify''. The red tags with strikethrough lines indicate the wrong predictions.}\label{fig:example}
\end{figure}

The second is \textbf{long term dependency}. Unlike pure POS tagging task which can utilize contextual features on word level, joint S\&T task usually extracts the contextual features on character level. Thus, the joint model need longer dependency on character level. As the example shown in Figure \ref{fig:example}, conditional random field (CRF) model makes mistakes on words ``reform'' and ``simplify'' since it is hard for CRF to disambiguate the POS tags without using long distance information. However, restricted by model complexity and data sparsity, a larger window size (greater than 5) will instead hurt the performance. Therefore, how to exploit the long distance information without increasing the model complexity is crucial to joint S\&T task.

In order to address these two problems, we propose a feature-enriched neural model for joint S\&T task, which consists of several key components: (1) a convolutional layer to simulate compositional features as complex hand-crafting features; (2) a pooling layer to select the most valuable features; (3) a bi-directional long short-term memory (BLSTM) layer on the top to carry long distance dependency information. In addition, we introduce a highway layer \cite{srivastava2015highway} to increase the depth of architecture and obtain more sophisticated feature representation without sufftering from the problem of gradient vanishing, leading to fast convergence.

Our contributions could be summaries as follows:
\begin{enumerate}
  \item We propose a customized neural architecture for joint S\&T task, in which each component is designed according to its specific requirements, instead of a general deep neural model.
  \item Our model can alleviate two crucial problems: model complexity and long term dependency in joint S\&T task.
  \item We evaluate our model on five different datasets. Experimental results show that our model achieves comparable performance to the previous sophisticated feature based models, and outperforms the previous neural models.
\end{enumerate}

\section{Neural Models for Joint S\&T}

\noindent The joint S\&T task is usually regraded as a character based sequence labeling problem.

In this paper, we employ the \{B\,M\,E\,S\} tag set $\T_{SEG}$ (indicating the Begin, Middle, End of the word, and a Single character word respectively) for word segmentation and the tag set $\T_{POS}$ (varies from dataset to dataset) for POS tagging.

The tag set $\T$ of our joint  S\&T task would be the cross-label set of $\T_{SEG}$ and $\T_{POS}$ as illustrated in Figure \ref{fig:example}.

\begin{figure}[t]
 \centering
 \hspace{2em}
 \includegraphics[width=0.38\textwidth]{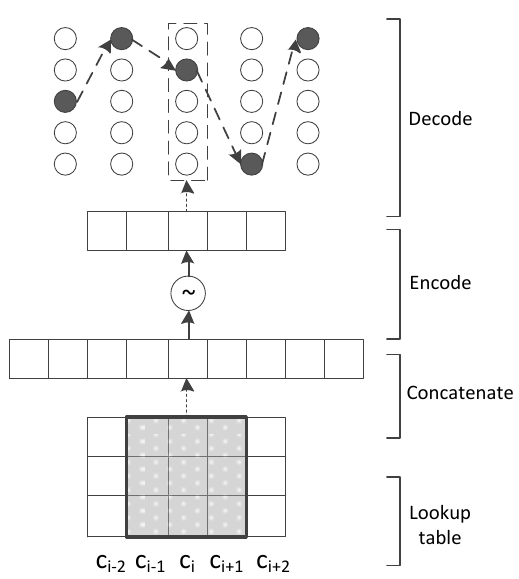}
 \caption{General neural network based architecture for joint S\&T. The solid arrow denotes that there is a weight matrix on the link, while the dashed one denotes none.}\label{fig:ConventionalModel}
\end{figure}

Conventional neural network based model for sequence labeling task usually consists of three phases. Figure \ref{fig:ConventionalModel} gives the illustration.
\begin{figure*}[t]
 \centering
 \includegraphics[width=0.95\textwidth]{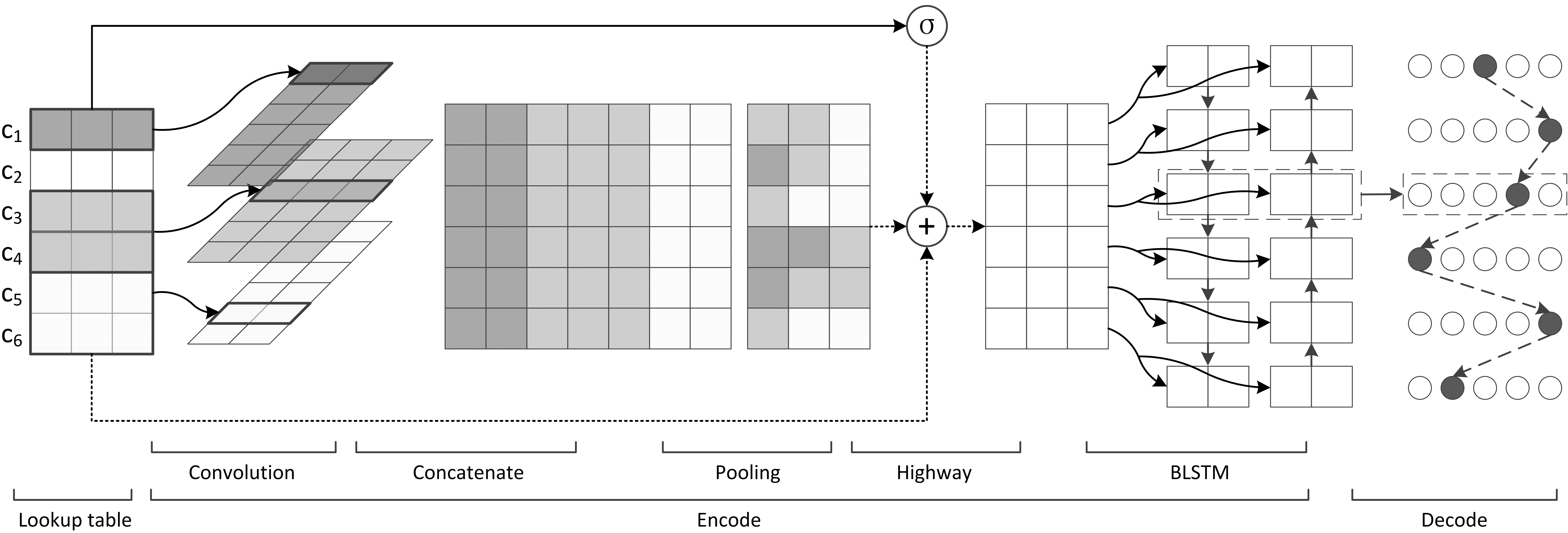}
 \caption{The proposed feature-enriched neural model for joint S\&T task. The solid arrow denotes that there is a weight matrix on the link, while the dashed one denotes none.}\label{fig:ProposedModel}
\end{figure*}
\subsection{Lookup Table Phase}
\noindent In order to represent characters as distributed vectors, we usually apply a feed-forward neural layer on the top of the one-hot character representations. The parameter matrix of the neural layer is called character embedding matrix $\mathbf{E} \in \R^{|C| \times d}$, where $C$ is the character set and $d$ is the dimensionality of the character embeddings. For a given sentence $c_{1:n}$ of length $n$, the first step is to lookup embeddings of the characters in the current window slide $c_{i-\left \lfloor\frac{k-1}{2}\right\rfloor:i+\left\lceil\frac{k-1}{2}\right\rceil}$ for the current character $c_i$ which is going to be tagged, where $k$ is a hyper-parameter indicating the window size. By concatenating the embeddings, we get the representation $\x_i$ for the current character $c_i$.
\subsection{Encoding Phase}
\noindent Usually, we apply a linear transformation followed by a non-linear function to the current input $\x_i$:
\begin{equation}
\h_i = \g({\bW_h}^\intercal \times \x_i + \bb_h),
\end{equation}
where $\bW_h \in \R^{kd \times h}$ and $\bb_h \in \R^{h}$ is the trainable parameters, and $h$ is the dimensionality of the hidden layer, $\g(\cdot)$ is a non-linear function which could be $\textbf{sigmoid}(\cdot)$, $\textbf{tanh}(\cdot)$, etc.

Then, we could get the score vector $\p_i \in \R^{|\T|}$ for each possible tags of current character $c_i$ by applying a linear transformation layer to the hidden layer $\h_i$:
\begin{equation}
\p_i = {\bW_p}^\intercal \times \h_i + \bb_p,
\end{equation}
where $\bW_p \in \R^{h \times |\T|}$ and $\bb_p \in \R^{|\T|}$ is the trainable parameters, and $\T$ is the joint tag set.

\subsection{Decoding Phase}
\noindent The decoding phase aims to select the best tag sequence $\hat{t}_{1:n}$, to maximize the reward function $\br(\cdot)$:
\begin{equation}
   \br(t_{1:n})= \sum_{i = 2}^n \left(\bA_{t_{i-1}t_i}\right)+\sum_{i = 1}^n \left(\p_i[t_i]\right),\label{eq:object}
\end{equation}
\begin{equation}
  \hat{t}_{1:n} = \argmax_{t_{1:n} \in \textbf{T}(c_{1:n})} \br(t_{1:n}),
\end{equation}
where $\bA \in \R^{|\T| \times |\T|}$ is the transition parameter, indicating how possible a label will transfer to another. $\textbf{T}(c_{1:n})$ indicates all possible tag sequences for sentence $c_{1:n}$.

Also, we employ the Viterbi algorithm \cite{forney1973viterbi} to decode the best tag sequence in polynomial time complexity.

\section{A Feature-Enriched Neural Model for Joint S\&T}
\noindent The simple neural model presented above achieves good results on the joint S\&T task. However, the simple neural model, who concatenates the embeddings of contextual characters as features, is not as strong as models based on the hand-crafted features. Thus, a simple shallow neural is insufficient to tackle with ambiguous cases which rely on more sophisticated feature combinations and long distance dependencies.

To deal with these issues, we propose a feature-enriched neural model for joint S\&T task, which consists of three different types of neural layers, stacked one by one: (1) Convolutional layer; (2) Highway layer; (3) Recurrent layer. Figure \ref{fig:ProposedModel} gives the illustration.

\subsection{Convolutional Layer}
\noindent The simple neural model is just to concatenate the embeddings of characters in a local context, which cannot simulate the carefully designed features in traditional models.

To better model the complex compositional features as conventional feature based models, we use convolution layer to separately model different n-gram features for each character. Thus the feature of each character is the concatenation of corresponding columns of all different feature map sets. Then we apply a $k$-max pooling layer to select the most significant signals.

Concretely, we model uni-gram, bi-gram, $\dots$, $Q$-gram features by generating feature map sets $\hat{\bz}^1$, $\hat{\bz}^2$, $\dots$, $\hat{\bz}^Q$ correspondingly. Formally, the $q$-gram feature map set $\hat{\bz}^q$ is:
\begin{equation}
  \hat{\bz}^q_i = \tanh ( {\bW^q_{cov}}^\intercal \times \x_{i-\left \lfloor\frac{q-1}{2}\right\rfloor:i+\left\lceil\frac{q-1}{2}\right\rceil} + \bb), i \in [1,n],
\end{equation}
where $\bW^q_{cov} \in \R^{qd \times l_q}$ is the convolutional filter for $q$-gram feature map set, and $\x_{i-\left \lfloor\frac{q-1}{2}\right\rfloor:i+\left\lceil\frac{q-1}{2}\right\rceil} \in \R^{qd}$ is the concatenation of embeddings of characters $c_{i-\left \lfloor\frac{q-1}{2}\right\rfloor:i+\left\lceil\frac{q-1}{2}\right\rceil}$. Here, $l_q$ is the number of feature maps in $q$-gram feature map set and $\bb \in \R^{l_q}$ is a bias parameter. For marginal cases, we use wide convolution strategy, which means we receive the sequence in the same length as input by padding zeros to the input.

Then, we would represent the original sentence by concatenation operation as $\bz \in \R^{n \times \sum_{q = 1}^{Q}l_q}$:
\begin{equation}
  \bz_i = \oplus_{q = 1}^{Q} \hat{\bz}_i,
\end{equation}
where operator $\oplus$ is the concatenation operation.

After taking the $k$-max pooling operation, the representation of original sentence would be $\hat{\mathbf{X}} \in \R^{n \times d} = {[\hat{\x}_1,\hat{\x}_2,\dots,\hat{\x}_n]}^\intercal$, where $\hat{\x}_i$ is:
\begin{equation}
  \hat{\x}_i = k \max \bz_i, k = d.
\end{equation}

Hence, after convolutional layer, we would represent the given input sequence $\mathbf{X} \in \R^{n \times d} = {[\x_1, \x_2, \dots, \x_n]}^\intercal$ as $\hat{\mathbf{X}} = Cov(\mathbf{X})$.

\subsection{Highway Layer}
\noindent Highway layer \cite{srivastava2015highway} aims to keep gradient in very deep neural network. 
By introducing highway layer, we could simulate more complex compositional features by increasing the depth of our architecture. In addition, highway layer speeds up convergence speed and alleviates the problem of gradient vanishing.

As described above, we would represent the input sequence as $\hat{\mathbf{X}} = Cov(\mathbf{X})$ after the convolutional layer. By additionally adding the highway layer, the representation of the input sequence would be $\hat{\mathbf{X}}$ as:
\begin{equation}
  \hat{\mathbf{X}} = Cov(\mathbf{X}) \odot T(\mathbf{X}) + \mathbf{X} \odot C(\mathbf{X}),
\end{equation}
where operator $\odot$ indicates the element-wise multiplication operation. The $T(\cdot)$ is the transform gate and $C(\cdot)$ is the carry gate. We adopt a simple version, where we set $C(\cdot) = 1 - T(\cdot)$. Transform gate $T(\cdot)$ could be formalized as:
\begin{equation}
  T(\mathbf{X}) = \sigma({\bW_T}^\intercal \times \mathbf{X} + \bb_T),
\end{equation}
where $\bW_T \in \R^{d \times d}$ and $\bb_T \in \R^{d}$ are trainable parameters. Here $\sigma$ is the $sigmoid$ function.

\subsection{Recurrent Layer}

\noindent In joint S\&T task, it usually relies on long distance dependency and sophisticated features to disambiguate lots of cases. Thus, a simple shallow neural model is insufficient to capture long distance information.

Inspired by recent works using long short-term memory (LSTM) \cite{hochreiter1997long} neural networks, we utilize LSTM to capture the long-term and short-term dependencies.
LSTM is an extension of the recurrent neural network (RNN) \cite{Elman:1990}, which aims to avoid the problems of gradient vanishing and explosion, and is very suitable to carry the long dependency information.

By further adding LSTM layer on the top of $\hat{\mathbf{X}} \in \R^{n \times d} = [\hat{\x}_1,\hat{\x}_2,\dots,\hat{\x}_n]$, we would represent sentence $c_{1:n}$ as ${\mathbf{H}} \in \R^{n \times h} = {\textbf{LSTM}}(\hat{\mathbf{X}}) = [{\mathbf{h}_1},{\mathbf{h}_2},\dots,{\mathbf{h}_n}]$.
Specifically, LSTM layer introduces memory cell $\mathbf{c} \in \R^{h}$ which controlled by input gate $\mathbf{i} \in \R^{h}$, forget gate $\mathbf{f} \in \R^{h}$ and output gate $\mathbf{o} \in \R^{h}$. Thus, each output ${\mathbf{h}_i} \in \R^{h}$ would be calculated as:
\begin{align}
\left[ \begin{array}{c}
            \mathbf{i}_i \\
            \mathbf{o}_i \\
            \mathbf{f}_i \\
            \hat{\mathbf{c}}_i
            \end{array}\right] &= \left[\begin{array}{c}
                                            \sigma\\
                                            \sigma\\
                                            \sigma\\
                                            \phi
                                        \end{array}\right]
                                                 \left( {{\bW}_g}^\intercal
                                                                \left[\begin{array}{c}
                                                                        \hat{\x}_i\\
                                                                        {\mathbf{h}}_{i-1}
                                                                \end{array}\right]
                                                        + {\bb}_g
                                                \right), \\
               \mathbf{c}_i    &= \mathbf{c}_{i - 1} \odot \mathbf{f}_i + \hat{\mathbf{c}}_i \odot \mathbf{i}_i, \\
               {\mathbf{h}}_i &= \mathbf{o}_i \odot \phi( \mathbf{c}_i ),
\end{align}
where ${\bW}_g \in \R^{4h \times (d + h)}$ and ${\bb}_g \in \R^{4h}$ are trainable parameters.
Here, the hyper-parameter $h$ is dimensionality of $\mathbf{i}$, $\mathbf{o}$, $\mathbf{f}$, $\mathbf{c}$ and $\mathbf{h}$.
$\sigma(\cdot)$ is $sigmoid$ function and $\phi(\cdot)$ is $tanh$ function.

\paragraph{BLSTM} We also employ the bi-directional LSTM (BLSTM) neural network. Specifically, each hidden state of BLSTM is formalized as:
\begin{equation}
  {\mathbf{h}}_i = \overrightarrow{{\mathbf{h}}}_i \oplus \overleftarrow{{\mathbf{h}}}_i,
\end{equation}
where operator $\oplus$ indicates concatenation operation. Here, $\overrightarrow{{\mathbf{h}}}_i$ and $\overleftarrow{{\mathbf{h}}}_i$ are hidden states of forward and backward LSTMs respectively.

\section{Training}

\noindent We employ max-margin criterion \cite{taskar2005learning} which provides
an alternative to probabilistic based methods by optimizing on the robustness of decision boundary directly.

In the decoding phase, if the predicted tag sequence for the $i$-th training sentence $c^{(i)}_{1:n_i}$ with the maximal score is ${\hat{t}}^{(i)}_{1:n_i}$:
\begin{equation}
  {\hat{t}}^{(i)}_{1:n_i} = \argmax_{t^{(i)}_{1:n_i} \in \textbf{T}(c^{(i)}_{1:n_i})} \br(t^{(i)}_{1:n_i}; \theta),
\end{equation}
the goal of the max-margin criterion is to maximize the score of the gold tag sequence ${t^*}^{(i)}_{1:n_i} = {\hat{t}}^{(i)}_{1:n_i}$ with a margin to any other possible tag sequence $t^{(i)}_{1:n_i} \in \textbf{T}(c^{(i)}_{1:n_i})$:
\begin{equation}
\br({t^*}^{(i)}_{1:n_i}; \theta) \geq \br(t^{(i)}_{1:n_i}; \theta) + \Delta({t^*}^{(i)}_{1:n_i}, t^{(i)}_{1:n_i}),
\end{equation}
\begin{equation}
  \Delta({t^*}^{(i)}_{1:n_i}, t^{(i)}_{1:n_i}) = \sum_{j = 1}^{n_i} \eta \textbf{1} \{{t^*}^{(i)}_j \neq t^{(i)}_j\},
\end{equation}
where $\Delta({t^*}^{(i)}_{1:n_i}, t^{(i)}_{1:n_i})$ is the margin function and hyper-parameter $\eta$ is a discount parameter.
Here, $\theta$ denotes all trainable parameters of our model.

Thus, the object is to minimize objective function $J(\theta)$ for $m$ training examples $(c^{(i)}_{1:n_i}, {t^*}^{(i)}_{1:n_i})_{i=1}^m$:
\begin{equation}
 J(\theta) = \frac{1}{m}\sum_{i=1}^{m}l_i(\theta) + \frac{\lambda}{2}\|\theta\|_2^2,
\end{equation}
\begin{align}
 l_i(\theta) =& \max_{t^{(i)}_{1:n_i} \in \textbf{T}(c^{(i)}_{1:n_i})}(\br(t^{(i)}_{1:n_i}; \theta) + \Delta({t^*}^{(i)}_{1:n_i}, t^{(i)}_{1:n_i})) \notag \\
 &-\br({t^*}^{(i)}_{1:n_i}; \theta).
\end{align}

\section{Experiments}
\subsection{Datasets}
\begin{table}\small\setlength{\tabcolsep}{3pt}
\centering
\begin{tabular}{|c|c|c|*{3}{r|}}
 \hline
 Datasets & Splits & Splits& AVG$_w$ & $N_{\text{sentence}}$&$|\mathcal{D}_W|$\\
 \hline
  \hline

  \multirow{2}*{CTB}&Train&\multirow{6}*{Sighan 2008}&27.4&23,444&42k\\
  &Test&&28.8&2,079&10k\\
  \cline{1-2}
    \cline{4-6}
  \multirow{2}*{PKU}&Train&&16.7&66,691&55k\\
  &Test&&24.3&6,424&18k\\
    \cline{1-2}
    \cline{4-6}
  \multirow{2}*{NCC}&Train&&28.4&18,869&45k\\
  &Test&&28.5&3,595&18k\\
  \hline
   \hline
  \multirow{3}*{CTB-5}&Train&1-270, 400-1151&27.3&18,086&37k\\
  &Dev&301-325&19.4&350&2k\\
  &Test&271-300&23.0&348&2k\\
  \hline
   \hline
     \multirow{2}*{CTB-7}&Train&1-4197&24.0&41,266&52k\\
  &Test&4198-4411&20.6&10,181&21k\\
  \hline
 \end{tabular}
\caption{Details of five datasets. $\mathcal{D}_W$ is the dictionary of distinct words. $N_{\text{sentence}}$ indicates the number of sentences. AVG$_w$ is the average word number in a sentence.}\label{tab:info_datasets}
\end{table}

\noindent We evaluate proposed architecture on five datasets: CTB, PKU, NCC, CTB-5, CTB-7. Table \ref{tab:info_datasets} gives the details of five datasets. We use the first 10\% data of shuffled train set as development set for CTB, PKU, NCC and CTB-7 datasets.
\begin{itemize}
  \item \textbf{CTB, PKU and NCC} datasets are from the POS tagging task of the Fourth International Chinese Language Processing Bakeoff \cite{moe2008fourth}.
  \item \textbf{CTB-5} dataset is the version of Penn Chinese Treebank 5.1, following the partition criterion of \cite{moe2008fourth,Jiang:2009,Sun:2012}
  \item \textbf{CTB-7} dataset is the version of Penn Chinese Treebank 7.0. It consists of different sources of documents (newswire, magazine articles, broadcast news, broadcast conversations, newsgroups and weblogs). Since the web blogs are very different with news texts, we try to evaluate the robustness of our model by testing on web blogs and training on the rest of dataset. 
\end{itemize}
\begin{table} \setlength{\tabcolsep}{3pt}
\centering
\begin{tabular}{|c|c|}
  \hline
  Window size &$k = 1$\\
  Character embedding size &$d = 50$\\
  Initial learning rate &$\alpha = 0.2$\\
  Margin loss discount &$\eta = 0.2$\\
  Regularization &$\lambda = 10^{-4}$\\
  LSTM dimensionality & $h = 100$\\
  Number of feature map sets &$Q = 5$\\
  Size of each feature map set $\hat{f}^q$&$l_{q=1}^Q = 100$\\
Batch size&$20$\\
  \hline
\end{tabular}
\caption{Hyper-parameter settings.}\label{tab:paramSet}
\end{table}
\begin{table*}[t]
\centering
\begin{tabular}{|c|*{9}{c|}}
 \hline
 \multirow{2}*{models} &
\multicolumn{3}{c|}{w/o LSTM} &  \multicolumn{3}{c|}{LSTM} & \multicolumn{3}{c|}{BLSTM} \\
 \cline{2-10}
  & P & R & F & P & R & F & P & R & F \\
 \hline
  w/o CNN	&-&-&-&89.24	&89.66	&89.45	&89.52	&89.74	&89.63\\
    CNN &88.24 &89.16		&88.70&89.35&89.71&89.53&89.75&89.61&89.68\\
  CNN+Pooling	&88.51 &89.00		&88.76&88.54	&89.13	&88.83	&88.91	&89.33	&89.12\\
 CNN+Pooling+Highway	&\textbf{90.14} &\textbf{90.34}		&\textbf{90.24}&\textbf{89.38}	&\textbf{89.73}	&\textbf{89.55}	& \textbf{90.23}	&\textbf{90.55}	&\textbf{90.39}	\\
  \hline
 \end{tabular}
\caption{Performances of different models on test set of CTB dataset.}\label{tab:model_selection}
\end{table*}

\begin{table*} [t]\setlength{\tabcolsep}{2pt}
\centering
\begin{tabular}{|c|*{9}{c|}|c|c|c||c|c|c|}
 \hline
 \multirow{2}*{models} &
  \multicolumn{3}{c|}{CTB} & \multicolumn{3}{c|}{PKU}& \multicolumn{3}{c||}{NCC}& \multicolumn{3}{c||}{CTB5}& \multicolumn{3}{c|}{CTB7} \\
 \cline{2-16}
  & P & R & F & P & R & F  & P & R & F & P & R & F& P & R & F \\
 \hline
    CRF	&\textbf{90.51}&90.23&90.37&90.00&89.12&89.56       &87.93&87.24&87.58&92.85&93.24&93.05&84.64&85.86&85.24\\

    \cite{qiu2013joint}	&89.11&89.16&89.13&89.41&88.58&88.99&-&-&-&\textbf{93.28}&93.35&\textbf{93.31}&-&-&-\\
    \hline
    \hline
MLP	&88.11&87.29&87.69&88.22&87.74&87.98&85.80&85.66&85.73&-&-&91.82$^*$&83.60&84.53&84.06\\
 Ours	&89.48&89.63&89.56&89.82&89.55&89.68&87.30&87.76&87.53&91.78&92.88&92.33&84.02&\textbf{86.26}&85.13\\
  Ours+Pre-train	&90.23&\textbf{90.55}&\textbf{90.39}&\textbf{90.27}&\textbf{90.05}&\textbf{90.16}&\textbf{88.37}&\textbf{89.16}&\textbf{88.76}& 92.88&\textbf{93.49}&93.19&\textbf{84.40}&86.25&\textbf{85.31}\\
  \hline
 \end{tabular}
\caption{Comparisons with previous models on test sets of CTB, PKU, NCC, CTB5 and CTB7 datasets.}\label{tab:res_sighan2008}
\end{table*}
\subsection{Hyper-parameters}
\noindent Table \ref{tab:paramSet} gives the details of hyper-parameter settings. Note that we set window size $k=1$ which means we only take the current character embedding into account instead of using window slice approach. According to experiment results, we find it is a tradeoff between model performance and efficiency to only use \{ uni-gram, bi-gram, $\dots$, $5$-gram \} convolutional feature map sets. Besides, we set sizes of all feature map sets consistently for simplicity. Following previous work \cite{pei2014maxmargin,chen2015long}, we also adopt bigram-character embedding in this paper.

\subsection{Effects of Components}
\noindent We experiment several models by using different neural component layers as shown in Table \ref{tab:model_selection}. The model incorporating convolutional layer, pooling layer, highway layer, and BLSTM layer, achieves the best performance on F1 score (90.39) on test set of CTB dataset. Therefore, we would like to compare our approach with other previous works using this topology.

Notably, the conventional model using window slice approach (Figure \ref{fig:ConventionalModel}) for joint S\&T task can be viewed as a special case of our model when we only adopt a singe convolutional layer.
\subsubsection{Pooling Layer and Highway Layer}
\noindent To evaluate the effectiveness of pooling layer and highway layer, we incrementally add pooling layer and highway layer on the top of convolutional layer. As shown in Table \ref{tab:model_selection}, by adding pooling layer, the performance decrease a little for the loss of information. However, we get the better performance on F1 score (90.24) by additionally adding highway layer. Although the performance does not benefit from pooling layer much, the pooling layer extracts the most important features and meets the consistent dimensionality requirement to add highway layer.
Intuitively, highway layer helps simulating more complex compositional features by increasing the depth of architecture.

\subsubsection{Long Short-Term Memory Layer}

\noindent In this work, we introduce (B)LSTM layer to carry the long distance dependency. To evaluate (B)LSTM layer, we experiment different models with and without (B)LSTM layer. As shown in Table \ref{tab:model_selection}, we could get a relatively high performance by using LSTM or BLSTM layer only, which shows the capability of (B)LSTM in modeling features and carrying long distance information.

By introducing convolutional layer and highway layer, we could further boost the performance which benefits from the feature modeling capability of convolutional layer and highway layer.

\begin{table*} [t]\setlength{\tabcolsep}{3pt} \small
\centering
 \includegraphics[width=0.88\textwidth]{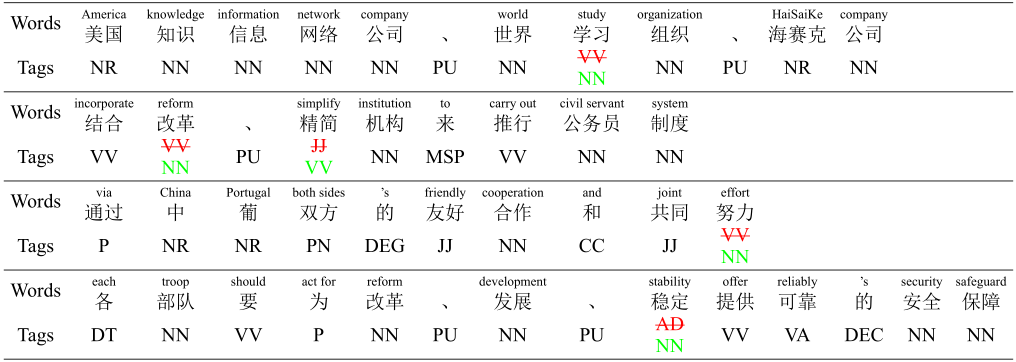}
\caption{Case study. The red tags with strikethrough lines indicate the wrong predictions, which are the results of the CRF model. The green tags are corrected predictions made by the proposed feature-enriched neural model. The black tags are correctly tagged by all models.}\label{tab:case_study}
\end{table*}

\subsection{Comparsion with Previous Works}

\noindent We compare proposed model with several previous works on five datasets on joint S\&T task. Experimental results are shown in Table \ref{tab:res_sighan2008}.

Conditional random field (CRF) \cite{Lafferty:2001} is one of the most prevalent and widely used models for sequence labeling tasks. \cite{qiu2013joint} aims to boost the performance by exploiting datasets with different annotation types. Multilayer perceptron (MLP) is our implementation of \cite{zheng2013deep}, a basic neural model for joint S\&T task. \cite{zheng2013deep} is a neural model which only use one layer of shallow feed forward neural network in the encoding phase. Our model indicates the model with convolutional layer, pooling layer, highway layer, and BLSTM layer. ``Pre-train'' indicates the pre-trained character embeddings which are trained on corresponding train set of each dataset using word2vec toolkit \cite{mikolov2013efficient}.

\renewcommand{\thefootnote}{\fnsymbol{footnote}}
\footnotetext{$^*$ \cite{zheng2013deep} only reported the results on CTB5 dataset for joint S\&T task.}
\renewcommand{\thefootnote}{\arabic{footnote}}
\paragraph{Result Discussion}
Our model outperforms the previous neural model on joint S\&T task and achieves the comparable performance with conventional hand-crafted feature based models.
As shown in Table \ref{tab:res_sighan2008}, compared to other previous methods, our model achieves the best performances on F1 scores (90.39, 90.16, 88.76, 85.31 on CTB, PKU, NCC and CTB-7 datasets respectively), and obtains comparable results on CTB5 dataset (93.19 on F1 score). As we know, the test set of CTB5 is very small so that previous work might overfit on that dataset. In addition, according to the experimental results, we find that the performance benefits a lot from pre-trained character embeddings. Intuitively, pre-trained embeddings give a more reasonable initialization for the non-convex optimization problem with huge parameter space.

Besides, the proposed model is quite efficient.
It only takes about half one hour per epoch using a small amount of memory (to train CTB) on a single GPU. Actually, it takes about ten hours to train our model (on CTB).

Experiments on CTB-7, whose train set and test set are on different domains, show the robustness of our model.
\subsection{Case Study}

\noindent We illustrate several cases from CTB-5 dataset. As shown in Table \ref{tab:case_study}, our approach performs well on cases with disambiguations which rely on long distance dependency. For instance, conditional random field (CRF) model makes mistakes on words ``reform'' and ``simplify'' since it is hard for CRF to disambiguate the POS tags without using the long distance (wider contextual) information.

\section{Related Works}

\noindent Recently, researches applied deep learning algorithms on various NLP tasks and achieved impressive results, such as chunking, POS tagging, named entity recognition for English \cite{collobert2011natural,tsuboi:2014:EMNLP2014,labeau-loser-allauzen:2015:EMNLP,ma2016end,santos2014learning,huang2015bidirectional}, and Chinese word segmentation and POS tagging for Chinese \cite{zheng2013deep,pei2014maxmargin,chen2017adversarial}.
These models learn features automatically which alleviate the efforts in feature engineering. However, joint S\&T is a more difficult task than Chinese word segmentation and POS tagging since it has a larger decoding space and need more contextual information and long distance dependency \cite{zhang2008joint,Jiang:2008,kruengkrai2009error,Zhang:2010,Sun:2011,qian2012joint,zheng2013deep,qiu2013joint,shen2014chinese}. Therefore, we need a customized architecture to alleviate these problems. In this work, we propose a feature-enriched neural model for joint S\&T task, and obtain great performance.

Besides, there are several similar neural models \cite{tsuboi:2014:EMNLP2014,labeau-loser-allauzen:2015:EMNLP,ma2016end,santos2014learning,huang2015bidirectional,kim2015character} . Instead of looking up word embedding table for each word in text, they tries to directly model English words by applying convolution layer on characters of words. Then they apply these word presentations to other tasks, such as POS tagging, name entity recognition, language modeling, etc. Unlike these models, we apply convolutional operation on sentence level, while they do within each word. Therefore they do not capture the features involving several words. Besides, we apply pooling operation along the feature size direction to get the most significant features.

\section{Conclusions}
\noindent In this paper, we propose a feature-enriched neural model for joint S\&T task, which better models compositional features and utilizes long distance dependency. Experimental results show that our proposed model outperforms the previous neural model and achieves comparable results with previous sophisticated feature based approaches.

\section*{Acknowledgments}
\noindent We would like to thank the anonymous reviewers for their valuable comments. This work is partially funded by National Natural Science Foundation of China (No. 61532011 and 61672162), Shanghai Municipal Science  and Technology Commission on (No. 16JC1420401).

\bibliography{acl2016}
\bibliographystyle{emnlp_natbib}

\end{document}